\documentclass[a4paper]{article}

\usepackage{INTERSPEECH2022}
\usepackage{graphicx}
\usepackage{amsmath}
\usepackage{amssymb}
\usepackage{tikz}
\usetikzlibrary{arrows.meta,positioning,calc,shapes.geometric}
\usepackage{multirow}
\usepackage{placeins}
\usepackage[colorlinks=true,citecolor=blue,linkcolor=blue,urlcolor=blue]{hyperref}

\title{Diarization-Guided Qwen-ASR Adaptation for Multilingual Two-Speaker Conversational Speech}
\name{Hao Wu$^{1,\ast}$, RongQi Han$^{1,\ast}$, Zhen Wang$^1$, Wei Liang$^2$, Wei Xu$^1$\thanks{$^\ast$Hao Wu and RongQi Han contributed equally to this work.}}
\address{
  $^1$Shanghai Qi Zhi Institute, Shanghai, China\\
  $^2$Megatronix (Beijing) Technology Co., Ltd}
\email{wuhao@sqz.ac.cn}

\begin{document}

\maketitle

\begin{abstract}
  This paper describes our self-designed system for Task 1 of the MLC-SLM 2026 Challenge for multilingual two-speaker conversational speech. The system combines a modular speaker diarization front end with a challenge-adapted Qwen3-ASR-1.7B recognizer. The diarization front end performs voice activity detection, subsegment generation, CAMPPlus speaker embedding extraction, two-speaker spectral clustering, and RTTM-based audio segmentation. The resulting speaker-attributed segments are grouped by language or region and decoded by the adapted ASR model. For ASR adaptation, we first perform supervised full fine-tuning on the official training data, then apply LoRA fine-tuning with synthetic speech generated by a three-pipeline TTS-based synthetic speech augmentation framework, and finally refine the model using GRPO reinforcement learning with rewards based on WER/CER and penalties for hallucination, repetition, and length deviation. On the official development set, the full system achieves an average tcpMER of 23.70, reducing the error rate by 6.83 absolute points relative to the released Qwen-ASR-1.7B performance. On the final evaluation set, the system achieves an average tcpMER of 17.97. Ablation results show that supervised fine-tuning provides the largest gain, while synthetic-speech LoRA adaptation and reinforcement learning further improve robustness.
\end{abstract}

\section{Introduction}

Transcribing two-speaker conversational recordings into speaker-attributed text is a demanding multilingual task: a single system must simultaneously decide who spoke when, segment the audio accordingly, and recognize what was said in each language. The difficulty is compounded by short turns, overlapping speech, frequent speaker switches, and the heterogeneous orthographies of the 21 language/region conditions covered by the MLC-SLM 2026 Challenge Task~1. Because diarization errors are inherited by the recognizer and recognition errors are inherited by the final score, treating speaker diarization (SD) and automatic speech recognition (ASR) as independent modules leaves substantial accuracy on the table; a carefully coupled SD+ASR pipeline is therefore central to a competitive system.

The challenge evaluates submissions with the time-constrained permutation mixed error rate (tcpMER), in which a language-adaptive mixed error rate (MER) is first computed per language---word error rate for space-separated scripts and character error rate for Japanese, Korean, and Thai---and then minimized over speaker permutations under time constraints, with diarization quality reported separately through the diarization error rate (DER). The 2026 Task~1 setting is defined by the official challenge page as multilingual conversational speech diarization and recognition \cite{mlcslm2026website}.

Our SD ASR system, denoted SQZ-Qwen-ASR-1.7B, targets Task~1 only. On the diarization side, we build a modular front-end within the 3D-Speaker framework \cite{chen20243dspeaker} that performs voice activity detection (VAD), CAMPPlus speaker embedding extraction, and spectral clustering with the speaker count fixed to two, followed by RTTM-driven audio cutting. On the recognition side, we start from the Qwen3-ASR-1.7B speech LLM \cite{shi2026qwen3asr} and adapt it to the MLC-SLM task setting through a three-stage recipe: full supervised fine-tuning (SFT) on the official training set, low-rank adaptation (LoRA) \cite{hu2022lora} on synthetic speech, and Group Relative Policy Optimization (GRPO) reinforcement learning that penalizes hallucinated and repetitive outputs. To feed the LoRA stage, we construct a three-pipeline TTS-based synthetic speech augmentation framework around the OmniVoice zero-shot voice cloning model \cite{zhu2026omnivoice}, producing 166{,}633 dialogue groups and 67{,}700 additional utterance-level samples that expand textual and speaker diversity. On the development set, SQZ-Qwen-ASR-1.7B reaches an average tcpMER of 23.70 in our local scoring pipeline, and on the evaluation set it achieves an average tcpMER of 17.97 according to the task leaderboard.

\section{System framework and training strategy}

\begin{figure*}[t]
  \centering
  \resizebox{0.82\textwidth}{!}{
  \begin{tikzpicture}[
    font=\footnotesize,
    >=Stealth,
    stageD/.style={draw=blue!55!black, fill=blue!20, rounded corners=3pt, line width=0.8pt, minimum width=2.35cm, minimum height=0.72cm, align=center},
    stageA/.style={draw=teal!55!black, fill=teal!20, rounded corners=3pt, line width=0.8pt, minimum width=2.35cm, minimum height=0.72cm, align=center},
    stageE/.style={draw=violet!55!black, fill=violet!18, rounded corners=3pt, line width=0.8pt, minimum width=2.35cm, minimum height=0.72cm, align=center},
    train/.style={draw=orange!70!black, fill=orange!25, rounded corners=3pt, line width=0.8pt, minimum width=2.35cm, minimum height=0.72cm, align=center},
    result/.style={draw=green!45!black, fill=green!22, rounded corners=3pt, line width=0.8pt, minimum width=2.35cm, minimum height=0.72cm, align=center},
    data/.style={draw=black!55, fill=white, rounded corners=3pt, line width=0.8pt, minimum width=2.35cm, minimum height=0.72cm, align=center},
    bandLabel/.style={align=center, font=\scriptsize\sffamily\bfseries, text=black!70, anchor=west},
    arrow/.style={->, line width=0.85pt, draw=black!65},
    dashedarrow/.style={->, dashed, line width=0.85pt, draw=black!55},
    flowLabel/.style={font=\scriptsize\sffamily, text=black!70, fill=white, inner sep=1.5pt}
  ]

    % ---- Swim-lane backgrounds ----
    \filldraw[fill=blue!5, draw=blue!20, rounded corners=6pt, line width=0.5pt] (-1.6, 2.1) rectangle (16.2, 3.4);
    \filldraw[fill=teal!5, draw=teal!22, rounded corners=6pt, line width=0.5pt] (-1.6, 0.75) rectangle (16.2, 2.0);
    \filldraw[fill=violet!5, draw=violet!22, rounded corners=6pt, line width=0.5pt] (-1.6, -0.6) rectangle (9.0, 0.7);
    \filldraw[fill=orange!5, draw=orange!22, rounded corners=6pt, line width=0.5pt] (-1.6, -3.7) rectangle (16.2, -1.05);

    % ---- Band labels (left margin) ----
    \node[bandLabel] at (-1.45, 2.75) {Diarization};
    \node[bandLabel] at (-1.45, 1.375) {ASR Inference};
    \node[bandLabel] at (-1.45, 0.05) {Evaluation};
    \node[bandLabel] at (-1.45, -2.375) {Model Adaptation};

    % ---- Row 1: diarization ----
    \node[data]    (wav)     at (2.2, 2.7)  {Evaluation\\wav};
    \node[stageD]  (vad)     at (4.9, 2.7)  {VAD\\subsegments};
    \node[stageD]  (emb)     at (7.6, 2.7)  {CAMPPlus\\embeddings};
    \node[stageD]  (cluster) at (10.3, 2.7) {Spectral\\clustering};
    \node[result]  (rttm)    at (13.0, 2.7) {RTTM\\diarization};

    % ---- Row 2: ASR inference (right-to-left) ----
    \node[stageA]  (splitwav) at (13.0, 1.35) {RTTM-based\\wav splitting};
    \node[stageA]  (list)     at (10.3, 1.35) {Raw list\\by language};
    \node[stageA]  (asr)      at (7.6, 1.35)  {Adapted Qwen-ASR\\batch inference};
    \node[result]  (stm)      at (4.9, 1.35)  {Hypothesis\\STM};

    % ---- Row 3: evaluation ----
    \node[stageE]  (norm)   at (4.9, 0)   {Text\\normalization};
    \node[stageE]  (metric) at (7.6, 0)   {tcpMER/DER\\evaluation};

    % ---- Row 4: model adaptation ----
    \node[data]    (official)  at (2.2, -1.85)  {Official\\training data};
    \node[train]   (sft)       at (4.9, -1.85)  {Full SFT\\Qwen-ASR};
    \node[train]   (lora)      at (7.6, -1.85)  {LoRA\\adaptation};
    \node[train]   (rl)        at (10.3, -1.85) {RL\\refinement};
    \node[data]    (synthetic) at (4.9, -3.0)   {LLM text\\+ TTS audio};
    \node[result]  (model)     at (10.3, -3.0)  {Adapted\\ASR model};

    % ---- Inference flow arrows ----
    \draw[arrow] (wav) -- (vad);
    \draw[arrow] (vad) -- (emb);
    \draw[arrow] (emb) -- (cluster);
    \draw[arrow] (cluster) -- (rttm);
    \draw[arrow] (rttm) -- (splitwav);
    \draw[arrow] (splitwav) -- (list);
    \draw[arrow] (list) -- (asr);
    \draw[arrow] (asr) -- (stm);
    \draw[arrow] (stm) -- (norm);
    \draw[arrow] (norm) -- (metric);

    % ---- Adaptation flow arrows ----
    \draw[arrow] (official) -- (sft);
    \draw[arrow] (synthetic) -- (lora);
    \draw[arrow] (sft) -- (lora);
    \draw[arrow] (lora) -- (rl);
    \draw[arrow] (rl) -- (model);

    % ---- Deploy: adapted model -> ASR inference (dashed, routed through free channel) ----
    \draw[dashedarrow] (model.east) -- (11.65,-3.0) -- node[flowLabel, pos=0.55] {deploy} (11.65,2.05) -- (7.6,2.05) -- (asr.north);

  \end{tikzpicture}
  }
  \caption{End-to-end framework of the SD ASR system, organized into four swim lanes. The \emph{Diarization} lane performs VAD, CAMPPlus speaker embedding extraction, spectral clustering, and RTTM generation. The \emph{ASR Inference} lane cuts audio according to RTTM, splits lists by language, and runs the adapted Qwen-ASR model to produce STM hypotheses. The \emph{Evaluation} lane normalizes text and computes tcpMER/DER. The \emph{Model Adaptation} lane trains Qwen-ASR via full SFT, synthetic-speech LoRA, and GRPO reinforcement learning; the resulting adapted model is deployed into the ASR inference lane (dashed arrow).}
  \label{fig:system_overview}
\end{figure*}
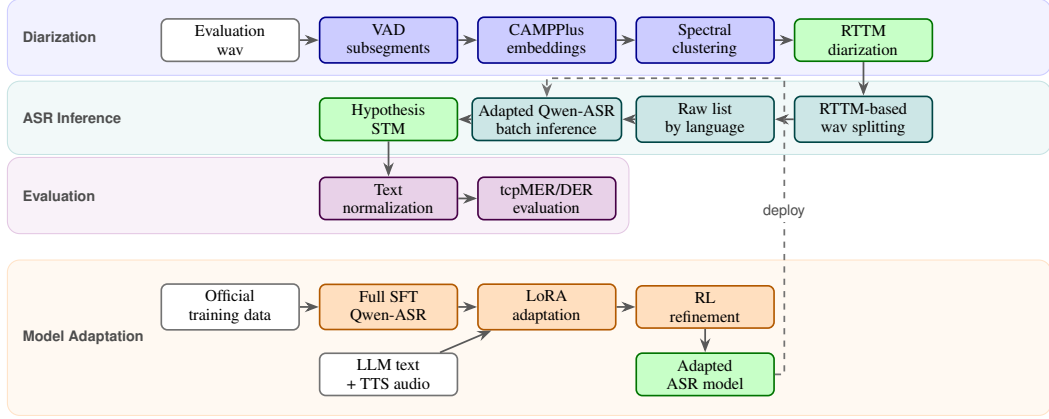

Figure~\ref{fig:system_overview} illustrates the overall framework of our system. The pipeline contains three functional parts: speaker diarization, audio segmentation and multilingual ASR, and text normalization with evaluation. The ASR model is trained separately and then integrated into the inference pipeline.

\subsection{Stage-wise inference pipeline}

The inference framework is organized as a multi-stage pipeline. It first performs voice activity detection with optional overlap refinement, then extracts speaker embeddings over short subsegments and clusters them into two-speaker turns. The resulting diarization drives a speaker-attributed segmentation of the audio, which is split by language or region and transcribed in batches by the adapted Qwen-ASR model. Diarization quality is monitored with DER, and the recognized text is normalized before end-to-end tcpMER scoring.

\subsection{Speaker diarization}

The diarization component is based on the 3D-Speaker framework \cite{chen20243dspeaker}. For each input recording, voice activity detection first identifies speech regions. The system then generates short subsegments and extracts CAMPPlus speaker embeddings, following the CAM++ speaker embedding architecture \cite{wang2023campp} and the production-oriented speaker embedding practice represented by Wespeaker \cite{wang2023wespeaker}. Spectral clustering assigns speaker labels, with the number of speakers fixed to two according to the challenge setting, and adjacent regions assigned to the same speaker are merged as a post-processing step. When enabled, overlap detection is used as an additional refinement stage. Concretely, VAD uses an FSMN-VAD model from FunASR \cite{gao2023funasr} (frame length 25~ms, frame shift 10~ms, chunk window 200~ms, 16~kHz) with a posterior decision threshold of 0.5 and SNR correction terms $-0.1/0.3$; CAMPPlus extracts 192-dimensional embeddings (80-dimensional fbank features) from subsegments of 1.5~s with a hop of 0.75~s (50\% overlap); spectral clustering fixes the cluster count to two (\texttt{min/max/oracle\_num\_spks}=2) with $\textit{mer\_cos}=0.8$, $\textit{pval}=0.012$, and $\textit{min\_cluster\_size}=4$, and adjacent same-speaker regions can be merged when the gap is below 0.8~s (this merge is disabled by default in our dev evaluation). This modular VAD--embedding--clustering design is a common alternative to end-to-end neural diarization \cite{fujita2019eend} and to separation-then-diarization pipelines used for meeting recognition \cite{vonneumann2024meeting}.

\subsection{Audio segmentation and ASR inference}

The ASR input is constructed from the diarization result. Each recording is cut according to its diarization regions, and long regions are further divided to keep the segment duration suitable for Qwen-ASR inference. The generated segments are grouped by language or region, covering 21 language-region conditions in the evaluation set. The adapted Qwen-ASR model is then used for batch transcription, producing speaker-attributed hypotheses for evaluation.

\subsection{Qwen-ASR adaptation}

The ASR model is adapted in multiple stages. First, Qwen-ASR-1.7B \cite{shi2026qwen3asr}, built on the Qwen audio-language modeling line \cite{chu2024qwen2audio,yang2024qwen2,yang2025qwen3}, is fully fine-tuned on the official MLC-SLM training set to learn the challenge transcription style. Second, task-related synthetic text is generated from the training context and converted into speech by a text-to-speech (TTS) system; the resulting synthetic speech pairs are used for LoRA fine-tuning \cite{hu2022lora}. Third, reinforcement learning is applied to improve recognition robustness and reduce unstable outputs from the language-model component. Full SFT uses AdamW with a linear schedule, warmup ratio 0.02, weight decay 0, and max gradient norm 1.0. It runs for 4 epochs at a peak learning rate of $2\!\times\!10^{-5}$, with a per-device batch size of 8, gradient accumulation of 4, and a maximum audio length of 30~s.

For LoRA fine-tuning, we insert adapters into the text decoder self-attention projections \texttt{q\_proj}, \texttt{k\_proj}, \texttt{v\_proj}, and \texttt{o\_proj} in all 28 layers (112 linear modules). We use rank $r=64$, $\alpha=128$, dropout 0.05. In addition to the LoRA adapters, we unfreeze only the audio projector weights and biases (\texttt{proj1}/\texttt{proj2}); all other base parameters remain frozen. Training runs for 5 epochs with peak learning rate $5\!\times\!10^{-5}$.

\subsection{GRPO-based reinforcement learning}

Unlike supervised fine-tuning that minimizes cross-entropy against a reference text, we cast ASR as sequential decision making and optimize Qwen3-ASR-1.7B with Group Relative Policy Optimization (GRPO), a critic-free policy optimization method introduced for language-model reinforcement learning \cite{shao2024deepseekmath}. For each speech input $x_i$ with reference $y_i$ and language label $l_i$, the policy $\pi_\theta$ samples $K=2$ candidate transcriptions $\hat{y}_{i,k}\sim\pi_\theta(\cdot\mid x_i,l_i)$, which are post-processed (prefix removal, punctuation/whitespace/case normalization) before reward computation.

The rule-based reward combines an accuracy term with stability penalties. The accuracy reward is $R_1=1-E(\hat{y},y)$, where $E(\cdot)$ auto-selects WER for space-separated languages and uses CER only for Thai, Japanese, and Korean, avoiding the distortion of applying WER to non-space-separated scripts. A hallucination rule strongly penalizes outputs that are over-long, low in unique-token ratio, or contain repeated $n$-grams or consecutive repeated tokens; repetition ($P_r$) and length-deviation ($P_l$) penalties are further applied. The final reward is
\begin{equation}
R=\mathrm{clip}(R_1-\lambda_r P_r-\lambda_l P_l,\,0,\,1),
\end{equation}
with accuracy weight $1.0$, repetition weight $\lambda_r=0.10$ (3-gram, $\geq2$ repeats, $\leq3$ consecutive), and length-deviation weight $\lambda_l=0.10$ (length-ratio tolerance 0.25); the hallucination penalty ($-1.0$) is returned directly when triggered by length ratio $>2.0$, unique-token ratio $<0.5$, or repeated 3-gram/consecutive limits. GRPO estimates the advantage from the $K$ group rewards without a critic, $A_{i,k}=(R_{i,k}-\mu_i)/(\sigma_i+\epsilon)$, where $\mu_i,\sigma_i$ are the group reward statistics; this token-level advantage drives a PPO-style clipped objective
\begin{equation}
\mathcal{L}_{\mathrm{PG}}=-\mathbb{E}_t\!\left[\min\big(\rho_t A_t,\,\mathrm{clip}(\rho_t,1{-}\epsilon,1{+}\epsilon)A_t\big)\right],
\end{equation}
with $\rho_t=\exp(\log\pi_\theta(a_t\mid s_t)-\log\pi_{\theta_{\mathrm{old}}}(a_t\mid s_t))$, augmented by a lightweight KL regularizer to the old policy for training stability. For rollout, the policy samples $K=2$ candidates per utterance with temperature 0.7, top-$k=20$, top-$p=0.95$, and a max response length of 256 . We use a PPO clip ratio $\epsilon=0.2$ ($\textit{clip\_ratio\_c}=3.0$), $\textit{ppo\_epochs}=1$, token-mean loss aggregation, and no entropy bonus; advantages follow the GRPO estimator normalized by group standard deviation with Monte-Carlo returns ($\gamma=\lambda=1$). Training uses AdamW with a constant learning rate of $1\!\times\!10^{-6}$ (no warmup, weight decay 0, gradient clip 1.0), an effective batch of 256 prompts ($\times K{=}2=512$ rollouts per step), and a max prompt length of 512. A low-variance KL loss regularizes the policy toward the reference with $\beta=0.005$; KL is not injected into the reward. This stage serves as the final ASR refinement after supervised fine-tuning and LoRA adaptation.

\begin{figure}[t]
  \centering
  \resizebox{\linewidth}{!}{
 \begin{tikzpicture}[
    font=\sffamily,
    >=Latex,
    model/.style={
        rounded corners=7pt,
        draw=teal!55!black,
        fill=teal!14,
        line width=0.9pt,
        minimum width=4.7cm,
        minimum height=1.2cm,
        align=center
    },
    reward/.style={
        rounded corners=7pt,
        draw=orange!70!black,
        fill=orange!16,
        line width=0.9pt,
        minimum width=3.1cm,
        minimum height=1.2cm,
        align=center
    },
    rewardDetail/.style={
        rounded corners=5pt,
        draw=orange!70!black,
        fill=orange!8,
        line width=0.7pt,
        minimum width=3.3cm,
        minimum height=1.5cm,
        align=center,
        font=\scriptsize\sffamily
    },
    inputNode/.style={
        rounded corners=5pt,
        draw=black!55,
        fill=black!5,
        line width=0.85pt,
        minimum width=3.5cm,
        minimum height=0.85cm,
        align=center
    },
    tokenBox/.style={
        draw=black!55,
        fill=white,
        line width=0.75pt,
        minimum height=0.6cm,
        inner sep=2pt
    },
    smallBox/.style={
        rounded corners=3pt,
        draw=black!60,
        fill=white,
        line width=0.8pt,
        minimum width=1.0cm,
        minimum height=0.6cm,
        align=center,
        font=\small\sffamily
    },
    lossNode/.style={
        rounded corners=6pt,
        draw=red!55!black,
        fill=red!12,
        line width=1.0pt,
        minimum width=2.6cm,
        minimum height=0.85cm,
        align=center,
        font=\small\sffamily\bfseries
    },
    legendBox/.style={
        rounded corners=4pt,
        draw=black!25,
        fill=black!3,
        line width=0.6pt
    },
    arrow/.style={->, line width=0.9pt, draw=black!70},
    thickArrow/.style={->, line width=1.4pt, draw=black!75},
    dashedArrow/.style={->, dashed, line width=0.8pt, draw=black!55},
    labelText/.style={font=\scriptsize\sffamily, text=black!70},
    flowLabel/.style={font=\scriptsize\sffamily, text=black!72, fill=white, inner sep=1.5pt}
]

% Token colors.
\definecolor{inputFill}{RGB}{255,246,214}
\definecolor{inputDraw}{RGB}{229,178,69}
\definecolor{outputFill}{RGB}{222,232,252}
\definecolor{outputDraw}{RGB}{134,164,218}
\definecolor{acousticFill}{RGB}{246,209,209}
\definecolor{acousticDraw}{RGB}{197,126,126}

% ---- Left pipeline: input -> prompt -> policy -> actor ----
\node[inputNode] (speechInput) at (-4.4,0) {Speech Audio + Text Prompt};
\node[tokenBox, minimum width=3.35cm] (promptTokens) at (-4.4,1.35) {};
\node[model] (policy) at (-4.4,3.0) {Qwen3-ASR-1.7B\\Policy Rollout\\{\scriptsize audio encoder inside}};
\node[model] (actor) at (-4.4,6.4) {Qwen3-ASR-1.7B\\Trainable Actor\\{\scriptsize audio encoder inside}};

% Prompt token strip: 2 text tokens + 3 audio tokens.
\foreach \x in {-1.30,-0.68} {
    \draw[draw=inputDraw, fill=inputFill, line width=0.9pt] ($(promptTokens.center)+(\x,0)$) circle (0.17);
}
\foreach \x in {-0.05,0.58,1.21} {
    \draw[draw=acousticDraw, fill=acousticFill, line width=0.9pt]
        ($(promptTokens.center)+(\x-0.17,-0.17)$) rectangle ++(0.34,0.34);
}

% ---- Candidate transcriptions (horizontal pair) ----
\node[tokenBox, minimum width=2.0cm] (candOne) at (-1.9,4.55) {};
\node[tokenBox, minimum width=2.3cm] (candTwo) at (0.55,4.55) {};
\node[labelText, fill=white, inner sep=1.0pt] at ($(candOne.north)+(0,0.26)$) {$o_1$};
\node[labelText, fill=white, inner sep=1.0pt] at ($(candTwo.north)+(0,0.26)$) {$o_2$};
\foreach \nodeName/\nTokens in {candOne/3,candTwo/4} {
    \foreach \i in {1,...,\nTokens} {
        \pgfmathsetmacro{\xoff}{0.48*(\i-(\nTokens+1)/2)}
        \draw[draw=outputDraw, fill=outputFill, line width=0.9pt]
            ($(\nodeName.center)+(\xoff,0)$) circle (0.17);
    }
}

% ---- Right side: reward detail + reward model ----
\node[rewardDetail] (rewardDetails) at (3.9,4.55) {Reward Components\\WER / CER\\Hallucination\\Repeat / Length};
\node[reward] (rewardModel) at (3.9,6.4) {Rule-based\\ASR Reward};

% ---- Top: policy probs, advantages, GRPO loss ----
\node[smallBox] (piOne) at (-5.4,7.85) {$\pi(o_1)$};
\node[smallBox] (piTwo) at (-3.4,7.85) {$\pi(o_2)$};
\node[smallBox] (advOne) at (2.9,7.85) {$A_1$};
\node[smallBox] (advTwo) at (4.9,7.85) {$A_2$};
\node[lossNode] (loss) at (-0.25,9.0) {GRPO Loss};

% ---- Arrows: left pipeline ----
\draw[arrow] (speechInput) -- (promptTokens);
\draw[arrow] (promptTokens) -- (policy);

% Sample arrow: policy -> candidates.
\draw[arrow] (policy.east) to[out=70,in=270] (candOne.south);
\node[flowLabel, anchor=north] at ($(candOne.south east)!0.55!(candTwo.south west)+(0,-0.08)$) {Sample $K{=}2$};

% Candidates -> actor (sampled trajectories reused in update).
\draw[thickArrow] ($(candOne.north east)+(-0.10,0.03)$) to[out=120,in=300] (actor.south east);

% Candidates -> reward detail.
\draw[arrow] (candTwo.east) -- (rewardDetails.west);

% Reward detail -> reward model.
\draw[arrow] (rewardDetails.north) -- (rewardModel.south);

% Actor -> policy probabilities for both sampled candidates.
\draw[arrow] ($(actor.north)+(-0.65,0)$) -- (piOne.south);
\draw[arrow] ($(actor.north)+(0.65,0)$) -- (piTwo.south);

% Reward model -> group-relative advantages for both sampled candidates.
\draw[arrow] ($(rewardModel.north)+(-0.45,0)$) -- (advOne.south);
\draw[arrow] ($(rewardModel.north)+(0.45,0)$) -- (advTwo.south);

% Each policy probability and advantage is passed to the GRPO loss.
\draw[arrow] (piOne.north) to[out=90,in=205] ($(loss.west)+(0,0.16)$);
\draw[arrow] (piTwo.north) to[out=90,in=190] ($(loss.west)+(0,-0.16)$);
\draw[arrow] (advOne.north) to[out=90,in=350] ($(loss.east)+(0,-0.16)$);
\draw[arrow] (advTwo.north) to[out=90,in=335] ($(loss.east)+(0,0.16)$);

% Actor update is driven by the loss (dashed feedback to emphasise training).
\draw[dashedArrow] (loss.south) to[out=270,in=90] node[flowLabel, pos=0.5] {update} (actor.north east);

% ---- Legend panel (bottom right) ----
\node[legendBox, minimum width=4.8cm, minimum height=2.6cm, anchor=north west] (legend) at (1.7,2.4) {};
\node[font=\scriptsize\sffamily\bfseries, text=black!70, anchor=north] at ($(legend.north)+(0,-0.15)$) {Legend};
\draw[draw=inputDraw, fill=inputFill, line width=0.9pt] (2.75,1.72) circle (0.15);
\node[anchor=west, labelText] at (3.15,1.72) {Input text token};
\draw[draw=outputDraw, fill=outputFill, line width=0.9pt] (2.75,1.22) circle (0.15);
\node[anchor=west, labelText] at (3.15,1.22) {Output text token};
\draw[draw=acousticDraw, fill=acousticFill, line width=0.9pt] (2.62,0.83) rectangle (2.88,0.55);
\node[anchor=west, labelText] at (3.15,0.69) {Speech audio token};
\node[anchor=center, labelText] at ($(legend.center)+(0,-0.98)$) {$A_k$: group-relative advantage};

\end{tikzpicture}
  }
  \caption{GRPO-based reinforcement learning for Qwen3-ASR. The speech audio and text prompt form the multimodal input; the policy model samples \(K=2\) candidate transcriptions, which are scored by rule-based ASR rewards (WER/CER, hallucination, repetition and length penalties). The resulting group-relative advantages \(A_k\) together with the policy probabilities \(\pi(o_k)\) drive the GRPO actor update.}
  \label{fig:grpo_rl}
\end{figure}
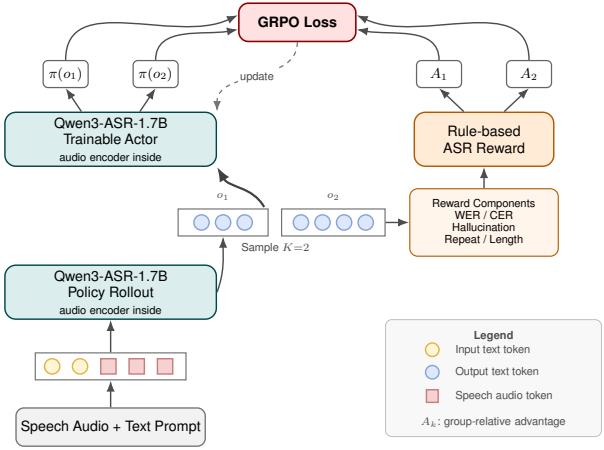

\section{Dataset}

\subsection{Training data}

The MLC-SLM 2026 Task~1 training set provides two-speaker conversational speech recorded at 16~kHz in quiet indoor environments, with oracle segmentation and speaker labels \cite{mlcslm2026website}. It covers 21 language/region conditions, built on the original 11-language release (English $\approx$500~h, other languages $\approx$100~h each, totaling $\approx$1{,}500~h) plus the 2026 additions of Tagalog, Urdu, Turkish, and the French(Canada), Spanish(Mexico), and Portuguese(Brazil) regional variants; the development set offers about 4~h per language under the same setting. We use the official training split for supervised fine-tuning, hold out the development set for model selection, and keep it as well as the evaluation set unseen. Transcripts follow the challenge text-normalization schema, applied identically in training and inference.

\subsection{TTS-based synthetic speech augmentation}

To expand lexical and acoustic coverage, we synthesize additional speech with OmniVoice, a multilingual zero-shot voice cloning model that synthesizes a target text using a reference speaker's timbre \cite{zhu2026omnivoice}. We design three complementary synthetic-speech generation pipelines (Table~\ref{tab:augmentation}), following the broader use of TTS-based synthetic speech augmentation for low-resource ASR \cite{zevallos2022tts,ibaraki2025frustratingly}. \emph{Textual expansion} synthesizes LLM-generated three-turn dialogue groups using real training utterances as reference (166{,}633 groups). Two further pipelines expand speaker and voice diversity: \emph{direct speaker injection} uses the training-set recordings as input to the first-stage-trained model to generate pseudo transcriptions, and only samples whose generated transcriptions differ from the original ground-truth labels are retained (32{,}700); and \emph{similarity-guided augmentation} retains recordings from public external datasets (e.g., Common Voice \cite{ardila2020commonvoice}) whose WeSpeaker \cite{wang2023wespeaker} speaker embeddings achieve a per-group cosine-similarity threshold of 0.76--0.83. The retained recordings are then used as references (35{,}000). In total, the framework contributes 234{,}333 synthesized training items when textual expansion is counted at the dialogue-group level; all items conform to the ASR manifest schema and are mixed into training without special handling. The external resource used for synthesis, OmniVoice, is disclosed per the challenge data-usage rules.

\begin{table}[t]
  \caption{Summary of TTS-based synthetic speech augmentation pipelines. Volume is counted as dialogue groups for textual expansion and as utterances for the two utterance-level synthesis strategies.}
  \label{tab:augmentation}
  \centering
  \footnotesize
  \setlength{\tabcolsep}{4pt}
  \begin{tabular}{p{2.05cm} p{2.15cm} p{2.35cm} r}
    \toprule
    \textbf{Strategy} & \textbf{Reference Source} & \textbf{Text Source} & \textbf{Volume} \\
    \midrule
    Textual Expansion        & Training-set recordings           & LLM-generated conversation    & 166{,}633 \\
    Direct Speaker Injection & Training-set recordings                & Train-set peusdo labels      & 32{,}700  \\
    Similarity-guided Aug.   & WeSpeaker-matched external recordings.     & Train-set true labels      & 35{,}000 \\
    % Error-targeted Synth.    & Train-set recordings  & ASR incorrect hypotheses      & 25{,}584  \\
    \midrule
    \textbf{Total}           & \multicolumn{3}{l}{\textbf{234{,}333 synthesized training items}} \\
    \bottomrule
  \end{tabular}
\end{table}

\begin{table*}[!t]
  \caption{Comparison results on the development set based in our local pipeline (SD: Speaker Diarization). All systems are evaluated with the time-constrained permutation mixed error rate (tcpMER), which uses tcpCER for Japanese, Korean, and Thai and tcpWER for the remaining languages. Baseline denotes the released Task~1 VibeVoice-ASR LoRA baseline \cite{mlcslm2026task1baseline}; Whisper-large-v3 and Omniasr-LLM-7B-v2 are official ASR baselines evaluated under the contest protocol; Qwen-ASR-1.7B is the released Qwen3-ASR-1.7B without fine-tuning; SQZ-Qwen-ASR-1.7B is our submitted Qwen3-ASR-1.7B checkpoint after full SFT, synthetic-speech LoRA, and reinforcement-learning refinement.}
  \label{tab:comparison}
  \centering
  \scriptsize
  \setlength{\tabcolsep}{2.6pt}
  \begin{tabular*}{\textwidth}{@{\extracolsep{\fill}}lccccc@{}}
    \toprule
    \textbf{Language} & \textbf{Baseline} & \textbf{Whisper-large-v3} & \textbf{Omniasr-LLM-7B-v2} & \textbf{Qwen-ASR-1.7B} & \textbf{SQZ-Qwen-ASR-1.7B} \\
    \midrule
    English-American   & 77.39 & 27.06 & 28.51 & 24.55 & 23.64 \\
    English-Australian & 81.50 & 19.53 & 21.22 & 16.45 & 14.12 \\
    English-British    & 67.60 & 26.03 & 27.97 & 24.32 & 22.99 \\
    English-Filipino   & 63.36 & 17.83 & 21.36 & 17.37 & 16.18 \\
    English-Indian     & 72.12 & 15.78 & 17.66 & 15.39 & 14.21 \\
    French-Canada      & 78.56 & 43.09 & 48.42 & 41.46 & 39.70 \\
    French             & 83.39 & 41.27 & 39.36 & 32.78 & 29.15 \\
    German             & 84.23 & 32.50 & 41.47 & 31.38 & 29.37 \\
    Italian            & 78.16 & 24.70 & 22.27 & 18.41 & 16.58 \\
    Japanese           & 81.46 & 41.71 & 41.92 & 30.43 & 29.21 \\
    Korean             & 81.33 & 31.58 & 33.25 & 27.17 & 25.14 \\
    Portuguese-Brazil  & 73.02 & 19.26 & 22.74 & 18.09 & 13.07 \\
    Portuguese         & 75.64 & 34.52 & 39.65 & 35.96 & 33.29 \\
    Russian            & 83.84 & 23.87 & 22.41 & 20.19 & 18.27 \\
    Spanish-Mexico     & 78.81 & 13.40 & 13.99 & 10.07 & 8.46  \\
    Spanish            & 82.51 & 21.31 & 21.30 & 18.95 & 17.76 \\
    Tagalog            & 81.09 & 32.99 & 46.75 & 49.49 & 33.90 \\
    Thai               & 83.67 & 21.91 & 19.91 & 16.46 & 14.38 \\
    Turkish            & 92.97 & 55.98 & 61.48 & 59.99 & 56.02 \\
    Urdu               & 89.63 & 44.36 & 34.18 & 102.84 & 19.77 \\
    Vietnamese         & 71.81 & 40.16 & 36.02 & 29.41 & 22.52 \\
    \midrule
    \textbf{tcpMER (Avg)} & \textbf{79.15} & \textbf{29.94} & \textbf{31.52} & \textbf{30.53} & \textbf{23.70} \\
    \bottomrule
  \end{tabular*}
\end{table*}

\begin{table}[!b]
  \caption{Ablation of the Task~1 system on the development set. Lower tcpMER is better. tcpMER values are averaged across the 21 language/dialect groups.}
  \label{tab:main_results}
  \centering
  \begin{tabular}{lc}
    \toprule
    \textbf{System} & \textbf{tcpMER} \\
    \midrule
    Base Qwen-ASR-1.7B          & 30.53 \\
    + Full SFT on official data & 24.3 \\
    + Synthetic-speech LoRA     & 23.78 \\
    + Reinforcement learning    & 23.70 \\
    \bottomrule
  \end{tabular}
\end{table}

\section{Experiments}

\subsection{Results}

We report tcpMER on the official development set under the challenge evaluation protocol. For ASR decoding we use greedy decoding at temperature~0 with a maximum output length of 256 tokens; diarization uses CAMPPlus embeddings and spectral clustering with the speaker count fixed to two, followed by RTTM-driven segmentation. The same diarization output is shared across all systems in Table~\ref{tab:comparison} so that differences reflect the ASR component only, and the sole external resource is the disclosed OmniVoice synthesizer used for augmentation.

Table~\ref{tab:comparison} compares our full system, SQZ-Qwen-ASR-1.7B, against the official SD ASR baseline and the released models such as Whisper-large-v3 under the same diarization-aware tcpMER. SQZ-Qwen-ASR-1.7B attains an average tcpMER of 23.70, compared with 79.15 for the official VibeVoice-ASR LoRA baseline, 30.53 for Qwen-ASR-1.7B, 29.94 for Whisper-large-v3, and 31.52 for Omniasr-LLM-7B-v2. Per-language gains over Qwen-ASR-1.7B are broad and largest where the base model is weakest (e.g., Urdu 102.84$\rightarrow$19.77, Tagalog 49.49$\rightarrow$33.90); Turkish remains the hardest case, where our system (56.02) is only on par with Whisper (55.98). Residual errors fall into diarization boundary and speaker-confusion mistakes on short overlapping turns, lexical errors on code-switching and named entities in low-resource languages, and punctuation or disfluency style mismatch. Although text-only LLM correction has been studied for ASR error correction \cite{ma2024asrec}, we leave it as future work since such correction can over-correct or hallucinate, especially for Japanese, Korean, and Thai.

\subsection{Ablation study}

Table~\ref{tab:main_results} isolates the contribution of each adaptation stage. Full supervised fine-tuning on the official data delivers the largest gain (30.53$\rightarrow$24.3) by aligning Qwen3-ASR-1.7B with the two-speaker conversational challenge setting and the official transcription conventions. Synthetic-speech LoRA gives a smaller, consistent further reduction to 23.78, with the biggest per-language impact on the hardest languages (Tagalog, Urdu) where the base model is weakest, by exposing the model to acoustic--linguistic confusions under-represented in the official data. Reinforcement learning adds a marginal average gain (23.78$\rightarrow$23.70); its main benefit is output stability---GRPO suppresses pathological repetitions and hallucinated tokens on long, noisy segments---at the cost of minor per-language regressions where the reward signal is noisier (e.g., Turkish). Overall, supervised fine-tuning drives the bulk of the improvement, synthetic speech broadens hard-language coverage, and RL acts as a low-risk stabilizer.

\FloatBarrier

\section{Conclusion}

We present an end-to-end speaker-diarization and ASR system for the MLC-SLM 2026 Task~1 challenge, built on Qwen3-ASR-1.7B and adapted through full supervised fine-tuning, synthetic-speech LoRA, and GRPO-based reinforcement learning. A three-pipeline TTS-based synthetic speech augmentation framework based on OmniVoice contributes 234{,}333 synthesized training items, counted as dialogue groups for textual expansion and as utterances for the two utterance-level synthesis strategies. On the official evaluation set, we get an average tcpMER of 17.97 according to the task leaderboard. On the development set, the full system achieves an average tcpMER of 23.70 on our local scoring pipeline, a 6.83-point improvement over the released Qwen-ASR-1.7B and substantially better than the Whisper-large-v3 and Omniasr-LLM-7B-v2 baselines. Ablations show that supervised fine-tuning provides the largest single gain, while synthetic-speech LoRA and RL add smaller, consistent refinements. Limitations remain in diarization robustness for similar-timbre speakers, synthetic-speech quality for the lowest-resource languages, and residual recognition instability on long noisy segments; future work will target stronger diarization--ASR coupling and more conservative post-hoc correction.

\bibliographystyle{IEEEtran}
\bibliography{mybib}

\end{document}